\documentclass[conference]{IEEEtran}
\IEEEoverridecommandlockouts

\usepackage{cite}
\usepackage{amsmath,amssymb,amsfonts}
\usepackage{algorithmic}
\usepackage{graphicx}
\usepackage{textcomp}
\usepackage{xcolor}
\usepackage{epsfig}
\usepackage{graphicx}

\usepackage{amsthm,bm}
\usepackage{graphics}
\usepackage{float}
\usepackage{array}
\usepackage{booktabs}
\usepackage{multirow}
\usepackage{lipsum}
\usepackage{hhline}
\usepackage{makecell}
\usepackage[numbers,sort,compress]{natbib}

\usepackage{caption}
\usepackage{subcaption}

\newcommand{\muv}{{\bm{\mu}}}
\newcommand{\Sigmam}{\bm{\Sigma}}

\newcommand{\zv}{\mathbf{z}}
\newcommand{\xv}{\mathbf{x}}

\newcommand{\Xm}{\mathbf{X}}
\newcommand{\Zm}{\mathbf{Z}}

\def\BibTeX{{\rm B\kern-.05em{\sc i\kern-.025em b}\kern-.08em
    T\kern-.1667em\lower.7ex\hbox{E}\kern-.125emX}}
\begin{document}

\title{Learning mappings onto regularized latent spaces for biometric authentication}

\author{
\IEEEauthorblockN{Matteo Testa, Arslan Ali, Tiziano Bianchi, Enrico Magli}
\IEEEauthorblockA{\textit{Department of Electronics and Telecommunications} \\
\textit{Politecnico di Torino, Italy}\\
Email: \{matteo.testa, arslan.ali, tiziano.bianchi, enrico.magli\}@polito.it}
}



\maketitle

\IEEEpubidadjcol

\begin{abstract}
   We propose a novel architecture for generic biometric authentication based on deep neural networks: RegNet. Differently from other methods, RegNet learns a mapping of the input biometric traits onto a target distribution in a well-behaved space in which users can be separated by means of simple and tunable boundaries. More specifically, authorized and unauthorized users are mapped onto two different and well behaved Gaussian distributions. The novel approach of learning the mapping instead of the boundaries further avoids the problem encountered in typical classifiers for which the learnt boundaries may be complex and difficult to analyze. RegNet achieves high performance in terms of security metrics such as Equal Error Rate (EER), False Acceptance Rate (FAR) and Genuine Acceptance Rate (GAR). The experiments we conducted on publicly available datasets of face and fingerprint confirm the effectiveness of the proposed system.
\end{abstract}


\section{Introduction}
\label{sec:intro}
Biometric authentication systems are becoming ubiquitous thanks to their undoubtful convenience as the users are authenticated based on information they inherently own, avoiding the need to remember passwords or provide keys.

Typically first a user's biometric traits, i.e. face, fingerprint or retina, are acquired through a sensor and then processed in such a way that discriminative features are extracted, and used to compute a template. This phase, which is referred to as ``enrollment", prepares the system to grant access only to the authorized user. In the subsequent phase, which is referred to as ``verification", the biometric trait of a user who requests to be granted access is given as input to the system. Therefore, in order to perform a decision, the stored template is matched with that of the fresh biometric trait through some appropriate distance metric. Based on the outcome of the matching procedure the user is authenticated or rejected. 

One of the most challenging aspects of a biometric authentication system is to have a negligible number of wrongly accepted users, yet providing excellent recognition performance by allowing the system to be invariant with respect to some transformations, e.g. pose and illumination.

With this work we propose to address this problem by means of deep neural networks.
Indeed, deep learning based methods have recently shown excellent performance at learning complex mappings \cite{goodfellow2014generative,almahairi2018augmented} and addressing difficult classification tasks \cite{krizhevsky2012imagenet}. In this regards, biometric authentication may be seen as a two-class classification problem in which the network has to learn how to correctly classify the enrolled user versus \textit{every other user}.

However, the downside of deep learning classification methods is that the boundaries which are learned in order to partition the feature space are highly complex and non-linear \cite{fawzi2017classification}. Works which addressed this specific problem \cite{fawzi2017classification,robustnessmagazine2017fawzi} concluded that most of the mass of the data points gathers close to the decision boundaries and as such this may strongly affect the robustness of the classifier.
Within the context of biometric authentication, this may lead two similar biometric traits of a user to be assigned to different classes, leading to an error.

To address this problem we propose a novel classification strategy in which the feature distributions are regularized so as to lead to simple boundaries between the classes, thereby reducing the probability of misclassification. 
In particular, we aim at designing a classifier having ``non-arbitrary" boundaries, which can be related to a clear data model and can eventually be tuned in order to achieve the desired performance. 
In order to reach our goal, we seek a compact yet meaningful mapping of the input biometric traits into a lower dimensional space which we will refer to as the latent space. 
Further, we constrain the latent space to be shaped in a simple and well-behaved manner (specifically, to follow Gaussian distributions) so that the region of the space corresponding to the authorized user is well-separated from that containing all the other users, i.e. we want the points to be separable with linear boundaries. 
The resulting system, which we will refer to as RegNet, employs simple threshold-based rules in this regularized latent space in order to discriminate between the authorized user and everyone else.

\subsection{Related work}

Several methods have been proposed to address the biometric authentication task when dealing with faces, fingerprints, retinas and gait. In this work we will focus on faces and fingerprints as we evaluate the performance of RegNet on these biometric traits.

Fingerprint authentication systems were among the first and most studied ones. Examples of non-learning based approached are \cite{chaoqiang2004hierarchical,ratha1996real} and \cite{sha2004orientation,abraham2011fingerprint} in which the matching is made on the global and local minutiae information respectively. Regarding deep learning based models, most of the effort has been put towards methods to extract and classify the minutiae. As an example, in \cite{jiang2016direct} a Convolutional Neural Network (CNN) is used to extract minutiae from raw fingerprint images. 

On the face authentication side, one of the earliest and most well-known examples is the Fisherfaces method \cite{swets1996using} which, thanks to the introduction of a supervised approach, improved robustness to illumination changes.
More recent approaches are based on low-dimensional representations of the faces; examples include sparse \cite{deng2012extended} and manifold \cite{he2005face} representations. A huge increase in performance has been obtained only recently with the advent of deep learning methods such as Facenet \cite{schroff2015facenet}, Deepface \cite{taigman2014deepface} and ArcFace \cite{deng2018arcface} which, by learning the features through deep CNN, are able to achieve state-of-the-art results.

From the above works it becomes clear that learning-based methods have mainly addressed biometric authentication as a classification problem.
To the best of our knowledge, this is the first approach to biometric authentication in which the paradigm is shifted from learning the classifcation boundaries to learning the mapping to a latent space. 
In this regard we mention some notable works which addressed the problem of learning a mapping into a regularized space. Examples include adversarial \cite{makhzani2015adversarial} and variational autoencoders \cite{kingma2013auto} in which the encoded representation is regularized to follow a target distribution. Regarding this latter work, we highlight that RegNet is conceptually different. In fact, as will become clear in the following, we propose a network which \textit{directly} generates samples from the intended distribution; this is in contrast to variational autoencoders in which the network learns the parameters of a distribution from which the samples are then drawn.

\section{Proposed Method}
\label{sec:format}

\begin{figure*}[tb]
    \centering
    \includegraphics[width=0.72\textwidth]{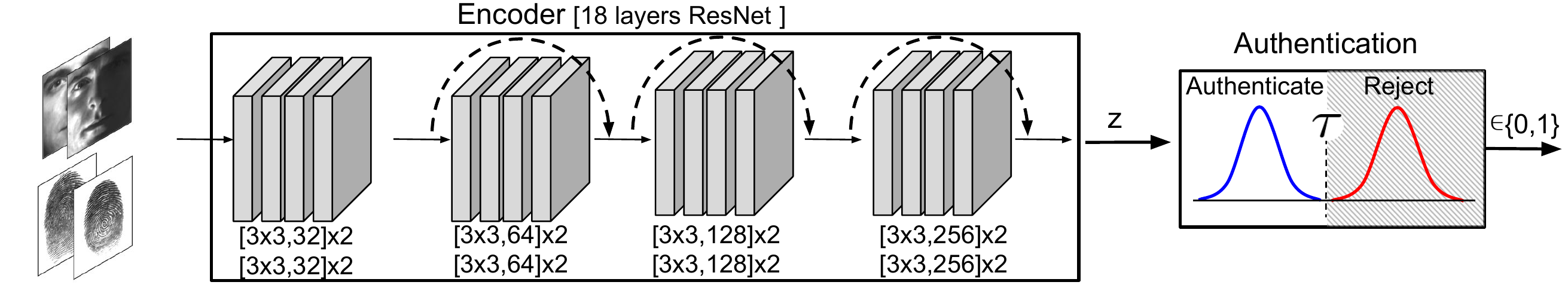}
    \caption[]{RegNet architecture. The biometric traits are given as input to the encoder; the output is a sample $\zv$ from either $\mathbb{P}_a$ or $\mathbb{P}_u$. During the authentication phase, given $\zv$ a thresholding decision can be applied to determine the user's class.}
    \label{fig:proposed_architecture}
    \vspace*{-0.5cm}
\end{figure*}

The main goal of RegNet is to learn a mapping from the distribution of the input biometric traits of authorized and unauthorized users to some well-behaved distributions in the latent space. 
More specifically, the biometric traits of the authorized user should be mapped to a target distribution whose mass center is far enough from that of the target distribution of unauthorized users. In this way, since that the latent space is well-behaved, a simple thresholding decision rule can be employed in order to discriminate among the two classes.

Being a biometric authentication system, RegNet operates in two phases: enrollment and authentication. In the following we will discuss in detail these two phases, which within the context of deep neural networks directly translate to the training and test phases.

\subsection{Enrollment}
In this phase the network has to learn the distribution of the biometric traits of the authorized user (respectively unauthorized users) and has to generate a sample drawn from the authorized (respectively unauthorized) target distribution.
It becomes evident that, in order to define a proper loss function, we should minimize a suitable distance metric between the distributions of the generated samples and the target ones. 
Thus, let us first define the desired target distributions $\mathbb{P}_a$ and $\mathbb{P}_u$ (for authorized and unauthorized users respectively) as two multivariate Gaussian distributions over a $d$-dimensional space:
\[
\mathbb{P}_a = \mathcal{N}(\muv_{Ta},\Sigmam_{Ta}),\;\mathbb{P}_u = \mathcal{N}(\muv_{Tu},\Sigmam_{Tu}),
\]
where $\Sigmam_{Ta} = \sigma^2_{Ta} \mathbb{I}_d$ and $\Sigmam_{Tu} =  \sigma^2_{Tu} \mathbb{I}_d$ are defined as diagonal covariance matrices and $\muv_{Ta} = \mu_{Ta} \mathbf{1}^T$, $\muv_{Tu} = \mu_{Tu} \mathbf{1}^T$ are the mean vectors.

At this point, in order to define a suitable distance metric let us define the output of the encoding network as $\zv = H(\xv)$, where $\zv \in \mathbb{R}^d$ is the latent mapping and $\xv \in \mathbb{R}^n$ is the input biometric trait. Further, let $\mathcal{B} = \{\mathcal{B}_{a=0},\mathcal{B}_{a=1}\}$ denote the set of all possible biometric traits and $a \in \{0,1\}$ an indicator variable such that $a=1$ represents the authorized user and $a=0$ represents all other unauthorized users. 
The goal is to learn an encoding function of the input biometric trait $\zv = H(\xv)$ such that $\zv \sim \mathbb{P}_a$ if $\xv \in \mathcal{B}_{a=1}$ and $\zv \sim \mathbb{P}_u$ if $\xv \in \mathcal{B}_{a=0}$, with $\mathbb{P}_a$ and $\mathbb{P}_u$ the target distributions in the latent space. 

We are now interested in computing the statistics of the generated samples $\zv$, thus we should recall that during the training the network is given as input a batch of biometric traits $\Xm \in \mathbb{R}^{b \times n}$ with $b$ being the batch size, thus resulting in $\Zm \in \mathbb{R}^{b \times d}$ after the encoding.
Therefore, we can compute the first and second order statistics (over a batch) of the encoded representations $\Zm_a,\Zm_b$ related to authorized ($\muv_{Oa}$,$\Sigmam_{Oa}$) and unauthorized ($\muv_{Ou}$,$\Sigmam_{Ou}$) input biometric traits respectively.
More specifically, we have that $\muv_{Oa}^{(i)} = \mathbb{E}[\Zm_a^{(i)}]$ and $\Sigmam_{Oa}^{(ii)} = \mathrm{var}(\Zm_a^{(i)})$, where $^{(i)}$ denotes the $i$-th colum and $^{(ii)}$ the $i$-th diagonal entry. 

Having defined the statistics of both target and encoded samples distributions, we can define a suitable metric to compare how far the distributions are from each other. More in detail we employ the KL divergence, which for multivariate Gaussian distributions (in case of authorized input biometric traits) can be written as:
\begin{align*}
    \mathcal{L}_a &= \frac{1}{2}\left[ \log \frac{|\Sigmam_{Ta}|}{|\Sigmam_{Oa}|} -d + \mathrm{tr}(\Sigmam_{Ta}^{-1}\Sigmam_{Oa}) \right.  +\\
    &+ \left. (\muv_{Ta}-\muv_{Oa})^\intercal \Sigmam_{Ta}^{-1}(\muv_{Ta}-\muv_{Oa}) \right]
\end{align*}
For the case of diagonal covariance matrices we are considering can be rewritten as
\begin{align*}
    \mathcal{L}_a = \frac{1}{2} \left[ \log\frac{\sigma_{Ta}^{2d}}{\prod_i \Sigmam_{Oa}^{(ii)}} -d + \frac{\sum_i \Sigmam_{Oa}^{(ii)}}{\sigma^2_{Ta}} + \frac{||\muv_{Ta} - \muv_{Oa}||_2}{\sigma^2_{Ta}} \right].
\end{align*}
In a similar fashion we can obtain $\mathcal{L}_u$ by considering the statistic's of both target and encoded distributions in the case of unauthorized input biometric traits.

Then, the loss function which the encoder network has to minimize is given by 
$\mathcal{L} = \frac{1}{2}\mathcal{L}_a + \frac{1}{2}\mathcal{L}_u$,
which achieves its minimum when the statistics of the two generated distributions will match that of the target ones. 

At this point we note that we are shaping the distribution of the encoded samples by only enforcing first and second order statistics. Indeed, from our experiments we have observed that these statistics are sufficient to shape the encoded samples distributions to closely follow the target ones. This leads us to conjecture that the encoder output tends to a maximum entropy distribution (Gaussian) and thus first and second order moments are sufficient to shape the latent space as intended.

\subsection{Authentication}
Following the enrollment phase, we can use the trained encoder to perform user authentication.

In the authentication phase the encoder network is used to compute the latent representation of the input biometric trait. Then, since the latent space is well-behaved, a threshold applied on the $\ell_2$ norm of the latent representation can be employed in order to output a decision.
Under the assumption of $\mu_{Ta} < \mu_{Tu}$ and $\sigma_{Ta} = \sigma_{Tu}$, the decision step can be formalized as follows:
\begin{align*}
    \begin{cases}
    \mathrm{accept} \; \mathrm{if} \; ||\zv||_2 \le \tau,\\
    \mathrm{reject} \; \mathrm{if} \ \  ||\zv||_2 > \tau,
    \end{cases}
    \label{eq:thresholding_function}
\end{align*}
where $\tau$ is an adjustable threshold that can be varied to obtain the desired trade-off between false acceptance rate (FAR) and false rejection rate (FRR).

\subsection{Architecture details}
RegNet deals with image biometric data, therefore we employ a convolutional neural network for the encoder architecture. More specifically, we use a ResNet-18 architecture \cite{resnet} made of four blocks, each of them consisting of an increasing number of $3\times 3$ filters, see Fig. \ref{fig:proposed_architecture}.
The last layer is a fully connected layer which maps the output of the last filter to $\zv$: the $d$-dimensional latent representation. For the experiments we set $d=3$ as it leads to better separation in the latent space and thus higher performance. Furthermore, we set $\mu_{Ta} = 0$, $\mu_{Tu} = 40$ and $\sigma_{Ta} = \sigma_{Tu} = 1$.
To optimize the network we employ Adam optimizer and use stochastic gradient descent over mini-batches of size $100$ samples.

\section{Experimental settings and results}
\label{sec:pagestyle}



\subsection{Datasets}
\label{sec:datasets}
In biometric authentication systems it is common to assume that the user puts him/herself in a controlled condition for the biometric acquisition (e.g. frontal face pose). This motivates us to consider constrained datasets, i.e. those dataset in which the biometric traits have been acquired in constrained conditions.

For the \textbf{face} authentication task we employ two commonly used datasets. The first dataset we employ is the CMU Multi-PIE dataset \cite{gross2010multi}. It consists of samples with different poses illumination and expressions. In total it has 750,000 samples of 337 subjects acquired in 4 different sessions. We consider the frontal poses of 129 subjects which are common in all 4 sessions. In total each user has 220 samples. We split the data as 75\% for training and the remaining 25\% for testing. For a single user enrollment out of 220 samples of authorized user, 165 are used for the training and remaining are left for testing. For unauthorized users enrollment out of 128 users, 96 users samples are used for the training and remaining 28 users samples are left for the testing.  We resize the images from 480x640x3 to 144x192x3.  In total, we consider 32 candidates for authorized user enrollment. Further to create more diverse samples, we employ the mixup strategy as described in \cite{zhang2017mixup}: positive and negative \textit{training} samples are mixed through convex combination. 

The second dataset we consider is the cropped version of \textit{extended Yale Face Database B} \cite{georghiades2001few}. It contains the frontal pose of $38$ subjects with varying illumination conditions with approximately $59$ samples of size $192 \times 168$ for every user.
For each enrollment, the dataset is split into training and test sets for both the authorized and unauthorized users. For each authorized user, out of the $59$ images, $49$ are used for training and $10$ for testing. For the case of unauthorized users, $31$ users are used for training ($1829$ samples) and $6$ are left for testing ($354$ samples).
The total number of training and test samples is $1878$ and $364$ respectively. Finally, by employing crops of size $184 \times 160$, the samples are augmented by an augmentation factor of $F = 81$ and $F_1 =25$. Further to create more diverse samples we employ mixup strategy as explained for CMU Multi-PIE dataset.

The \textbf{fingerprint} authentication experiments are performed on \textit{Fingerprint Verification Competition (FVC 2006) DB2} \cite{cappelli2007fingerprint} dataset. The dataset is made of $150$ users, each of them containing $12$ image samples acquired by an optical sensor. The images of size $560 \times 400$ are resized to $202 \times 149$.
For each enrollment, out of the $12$ images of each authorized user, $10$ are used for training and the remaining ones for testing. For the case of unauthorized users, $124$ users are used for training ($1488$ samples) and $25$ are left for testing ($300$ samples).
The total number of training and test samples is $1498$ and $302$ respectively. The dataset is augmented to augmentation factors of $F=289$ and $F_1=25$ by cropping the images down to $186 \times 133$ pixels. 
As done for the face dataset we employ mixup \cite{zhang2017mixup}.

\begin{figure*}[h]
\centering
\begin{subfigure}{.18\textwidth}
  \centering
  \includegraphics[width=\linewidth]{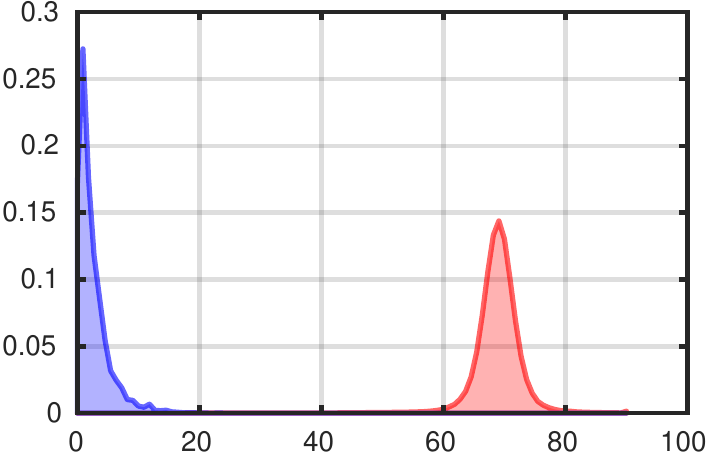}
  \caption[]{{RegNet}}
  \label{fig:sub2}
\end{subfigure}
\begin{subfigure}{.18\textwidth}
  \centering
  \includegraphics[width=\linewidth]{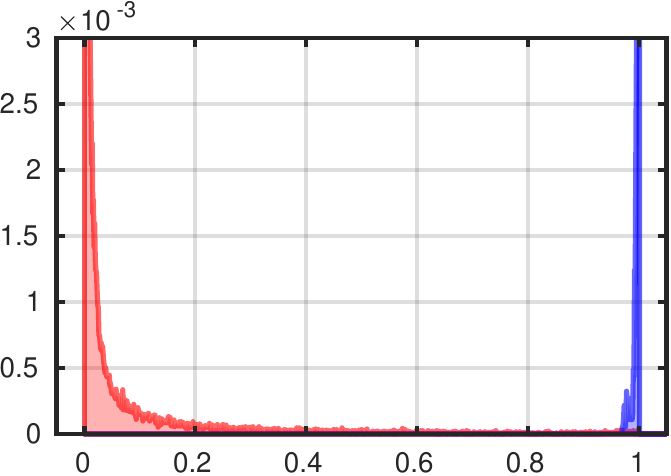}
  \caption[]{RegNet encoder class.}
  \label{fig:sub2}
\end{subfigure}
\begin{subfigure}{.18\textwidth}
  \centering
  \includegraphics[width=\linewidth]{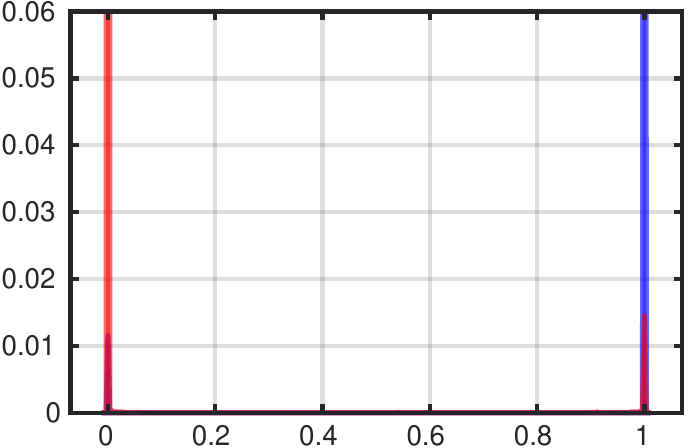}
  \caption[]{FaceNet + classifier}
  \label{fig:sub2}
\end{subfigure}
\begin{subfigure}{.18\textwidth}
  \centering
  \includegraphics[width=\linewidth]{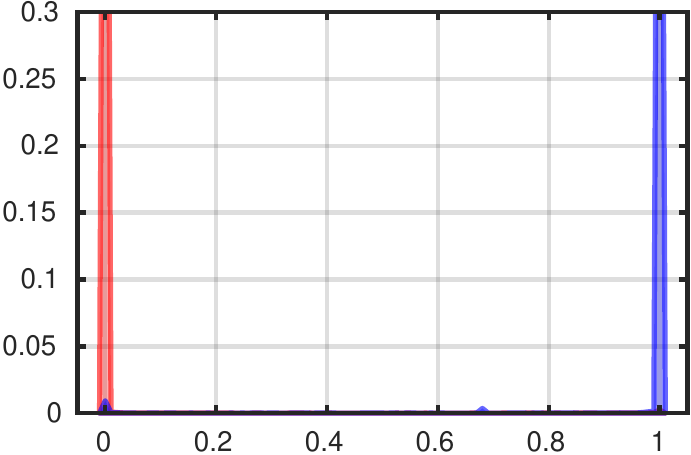}
  \caption[]{ArcFace + classifier}
  \label{fig:sub2}
\end{subfigure}
\begin{subfigure}{.18\textwidth}
  \centering
  \includegraphics[width=\linewidth]{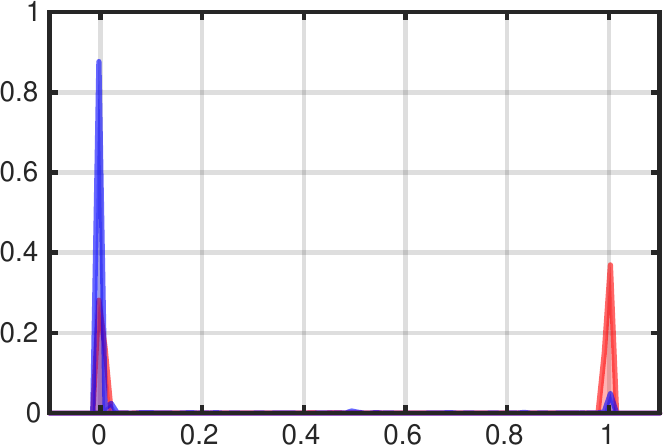}
  \caption[]{Fisherfaces}
  \label{fig:sub2}
\end{subfigure}
\caption[]{Face authentication scores for authorized users (blue) and unauthorized users (red) for Yale B. (a) Histogram of $||\zv||_2$ decision statistics of RegNet; (b) Histogram of the sigmoid outputs of RegNet encoder classifier; (c) Histogram of the sigmoid outputs of FaceNet embeddings classifier; (d) Histogram of the sigmoid outputs of ArcFace embeddings classifier; (e) Histogram of the normalized matching distances of Fisherfaces. The plots in (b)-(c) depict a detailed view to better appreciate the leakage effects.}
\vspace*{-0.2cm}
\label{fig:hist_faces_yale}
\end{figure*}
\begin{figure*}[h]
\centering
\begin{subfigure}{.18\textwidth}
  \centering
  \includegraphics[width=\linewidth]{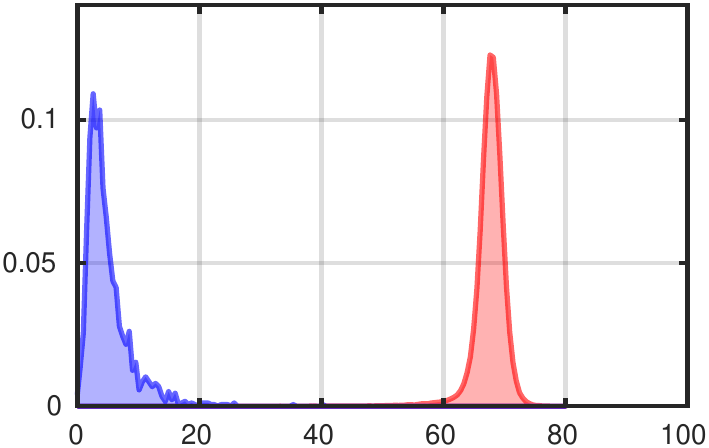}
  \caption[]{{RegNet}}
  \label{fig:sub2}
\end{subfigure}
\begin{subfigure}{.18\textwidth}
  \centering
  \includegraphics[width=\linewidth]{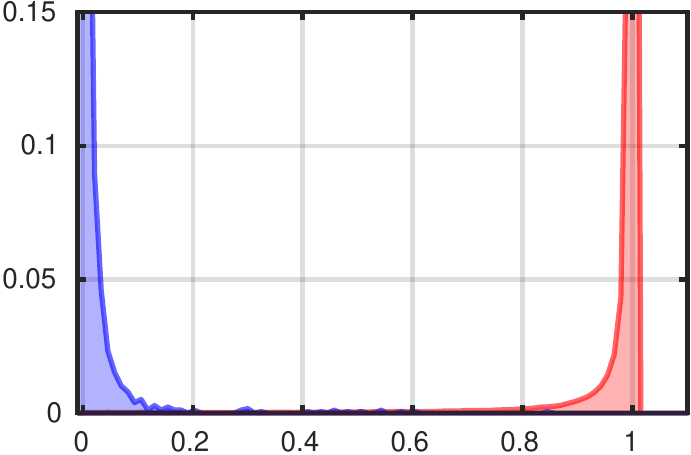}
  \caption[]{RegNet encoder class.}
  \label{fig:sub2}
\end{subfigure}
\begin{subfigure}{.18\textwidth}
  \centering
  \includegraphics[width=\linewidth]{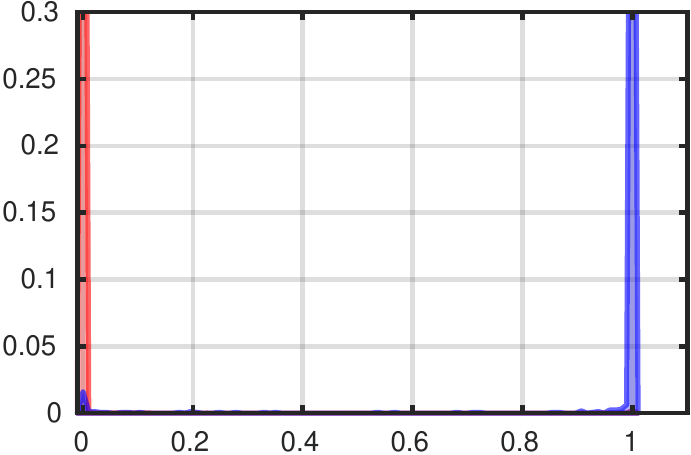}
  \caption[]{FaceNet + classifier}
  \label{fig:sub2}
\end{subfigure}
\begin{subfigure}{.18\textwidth}
  \centering
  \includegraphics[width=\linewidth]{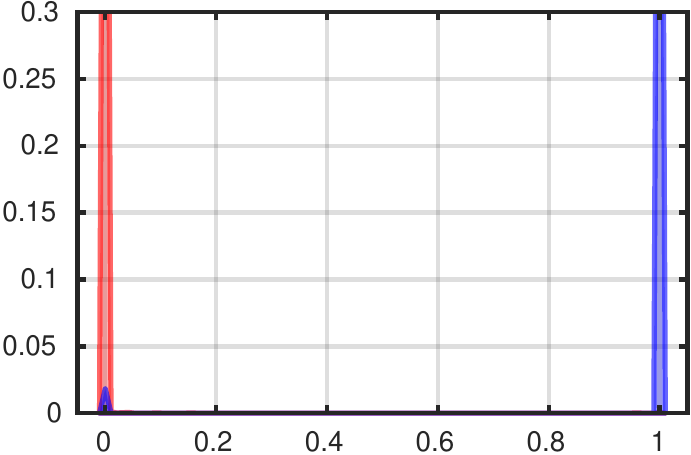}
  \caption[]{ArcFace + classifier}
  \label{fig:sub2}
\end{subfigure}
\begin{subfigure}{.18\textwidth}
  \centering
  \includegraphics[width=\linewidth]{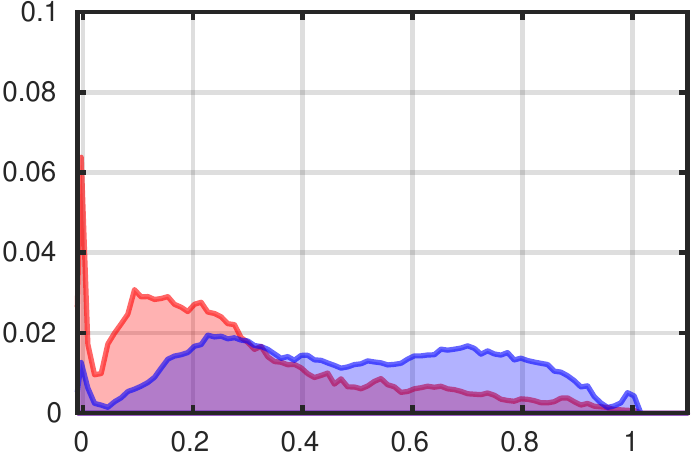}
  \caption[]{Fisherfaces}
  \label{fig:sub2}
\end{subfigure}

\caption[]{Face authentication scores for authorized users (blue) and unauthorized users (red) for Multi-PIE. (a) Histogram of $||\zv||_2$ decision statistics of RegNet; (b) Histogram of the sigmoid outputs of RegNet encoder classifier; (c) Histogram of the sigmoid outputs of FaceNet embeddings classifier; (d) Histogram of the sigmoid outputs of ArcFace embeddings classifier; (e) Histogram of the normalized matching distances of Fisherfaces. The plots in (b)-(c) depict a detailed view to better appreciate the leakage effects.}
\vspace*{-0.2cm}
\label{fig:hist_faces_multipie}
\end{figure*}

\begin{figure*}[h!]
\centering
\begin{subfigure}{0.2\textwidth}
  \centering
  \includegraphics[width=\linewidth]{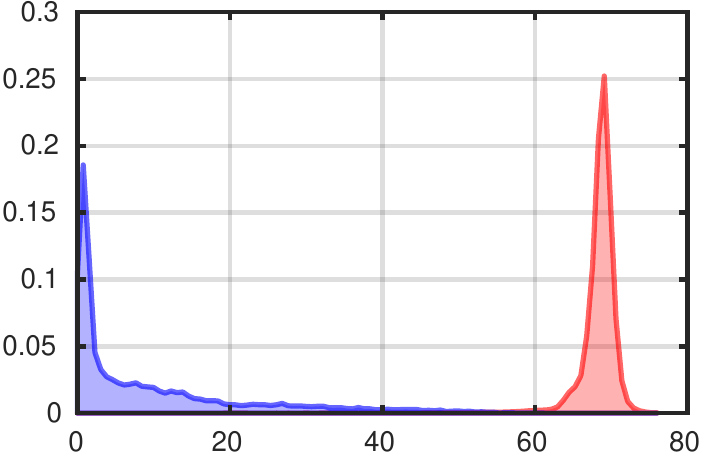}
  \caption[]{{RegNet}}
  \label{fig:RegNetfinger}
\end{subfigure}
\begin{subfigure}{.2\textwidth}
  \centering
  \includegraphics[width=\linewidth]{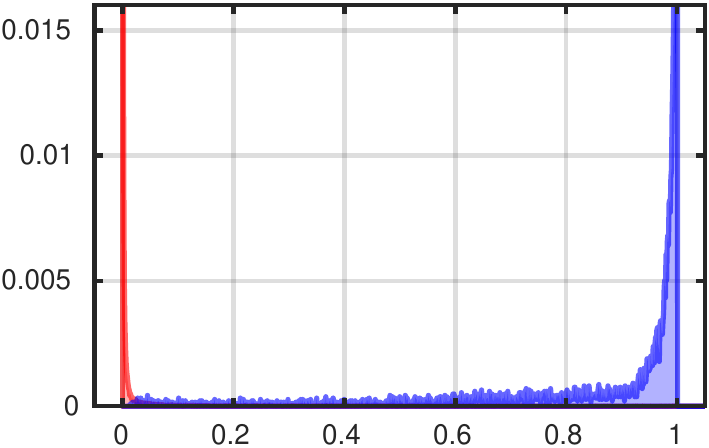}
  \caption[]{RegNet encoder class.}
  \label{fig:RegNetencoderfinger}
\end{subfigure}
\begin{subfigure}{.2\textwidth}
  \centering
  \includegraphics[width=\linewidth]{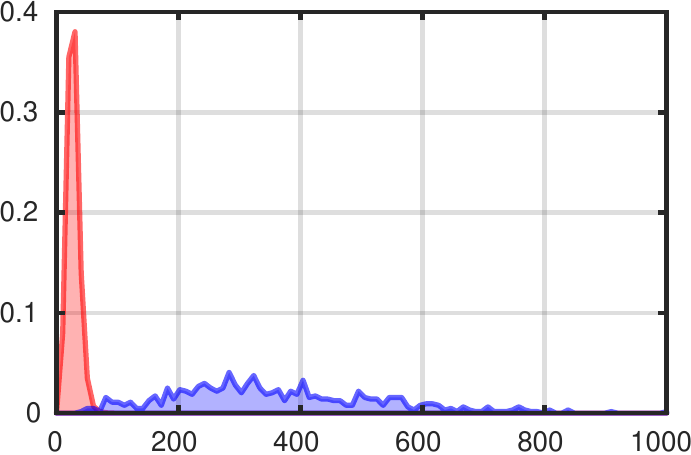}
  \caption[]{VeriFinger \cite{verifingerneuro}}
  \label{fig:verifinger}
\end{subfigure}
\begin{subfigure}{.2\textwidth}
  \centering
  \includegraphics[width=\linewidth]{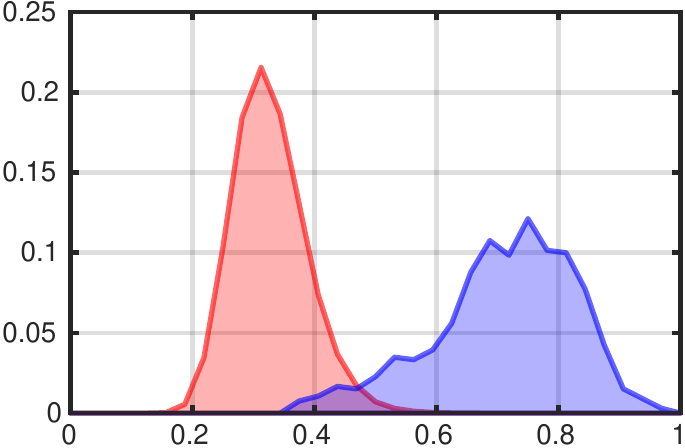}
  \caption[]{Hybrid approach \cite{abraham2011fingerprint}}
  \label{fig:sub2}
\end{subfigure}
\caption[]{Fingerprint authentication scores for authorized users (blue) and unauthorized users (red). (a) Histogram of $||\zv||_2$ decision statistics for RegNet; (b) Histogram of the sigmoid outputs of the RegNet encoder classifier; (c) histogram of the matching scores of Verifinger; (d) histogram of the matching scores of the hybrid approach. The plot in (b) depicts a detailed view to better appreciate the leakage effects.}
\label{fig:hist_fingerprints}
\vspace*{-0.3cm}
\end{figure*}

\begin{table*}[h]
 \centering
\resizebox{0.9\textwidth}{!}{
\begin{tabular}{ c c c c c c c c}
\hline
\textbf{Dataset}    & \textbf{Method}     & \textbf{EER$\%$}     & \textbf{GAR@$\mathbf{10^{-1}}$FAR$\%$}  &\textbf{GAR@$\mathbf{10^{-2}}$FAR$\%$} &\textbf{Accuracy@EER}   \\ \Xhline{3\arrayrulewidth}

 & {\color[HTML]{000000} \textbf{RegNet}} & {\color[HTML]{000000} \textbf{0.023 }}   &{\color[HTML]{009901} } \textbf{100.0}  & {\color[HTML]{009901} } \textbf{100.0} & {\color[HTML]{009901} } \textbf{99.977} \\ \cline{2-6} 
 
 & RegNet enc. classifier   & 0.040 &     100.0 & 100.0 & 99.960\\ \cline{2-6} 
  & FaceNet    & 1.286   &  98.819 & 98.712 & 98.714 \\
   \cline{2-6} 
  & ArcFace    & 0.893 &  99.159 & 99.108 & 99.107 \\
  \cline{2-6} 
 \multirow{-4}{*}{\textbf{Face - Yale B}}    

 & Fisherfaces & 15.351 &  84.215 &  61.135 & 84.649\\ \Xhline{2\arrayrulewidth}
  & 
 
 {\color[HTML]{000000} \textbf{RegNet}} & {\color[HTML]{000000} \textbf{0.045}}   & {\color[HTML]{000000} {\textbf{100.0}}}  & {\color[HTML]{000000} {\textbf{100.0}} }  & \textbf{99.955}  \\ \cline{2-6} 
 
 & RegNet enc. classifier     & 0.676       &  100.0 & 99.432  & 99.324  \\ \cline{2-6} 
 & FaceNet      & 0.930   &   99.368   &   99.201 & 99.070
\\ \cline{2-6} 
 & ArcFace      & 1.811  &   98.811   &   98.125 & 98.189
 \\ \cline{2-6} 

\multirow{-4}{*}{\textbf{Face - Multi-PIE}} 

& Fisherfaces  & 32.620   & 10.002 &  2.800 & 67.379 \\ \Xhline{3\arrayrulewidth}
\end{tabular}}
\caption[]{Performance comparison of RegNet with respect to other biometric authentication schemes for faces.}
\label{table:EER_faces}
\vspace*{-0.2cm}
\end{table*}

\begin{table*}[h]
 \centering
\resizebox{0.9\textwidth}{!}{
\begin{tabular}{ c c c c c c }
\hline
\textbf{Dataset}    & \textbf{Method}     & \textbf{EER$\%$}     & \textbf{GAR@$\mathbf{10^{-1}}$FAR$\%$}  &\textbf{GAR@$\mathbf{10^{-2}}$FAR$\%$} & \textbf{Accuracy@EER}    \\ \Xhline{3\arrayrulewidth} &
 {\color[HTML]{000000} \textbf{RegNet}} & {\color[HTML]{000000} \textbf{0.476 }}   & {\color[HTML]{000000} {\textbf{100.0 }}}  & {\color[HTML]{000000} {\textbf{99.934}} }  & \textbf{99.524} \\ \cline{2-6} 
 
 & RegNet enc. classifier     & 0.565      &  100.0 & 99.845 & 99.435 \\ \cline{2-6} 
 & Verifinger      & 0.758    &   100.0    &   99.796  & 99.361  \\ \cline{2-6} 

\multirow{-4}{*}{\textbf{Fingerprint FVC 2006}} 

& Hybrid approach \cite{abraham2011fingerprint}  & 3.200  &  98.182 &  94.854 & 96.799\\ \Xhline{3\arrayrulewidth}
\end{tabular}}
\caption[]{Performance comparison of RegNet with respect to other biometric authentication schemes for fingerprints.}
\label{table:EER_fingerprints}
\vspace*{-0.2cm}
\end{table*}

\begin{figure*}[tb]
\centering
\begin{subfigure}{.32\textwidth}
  \centering
  \includegraphics[width=\linewidth]{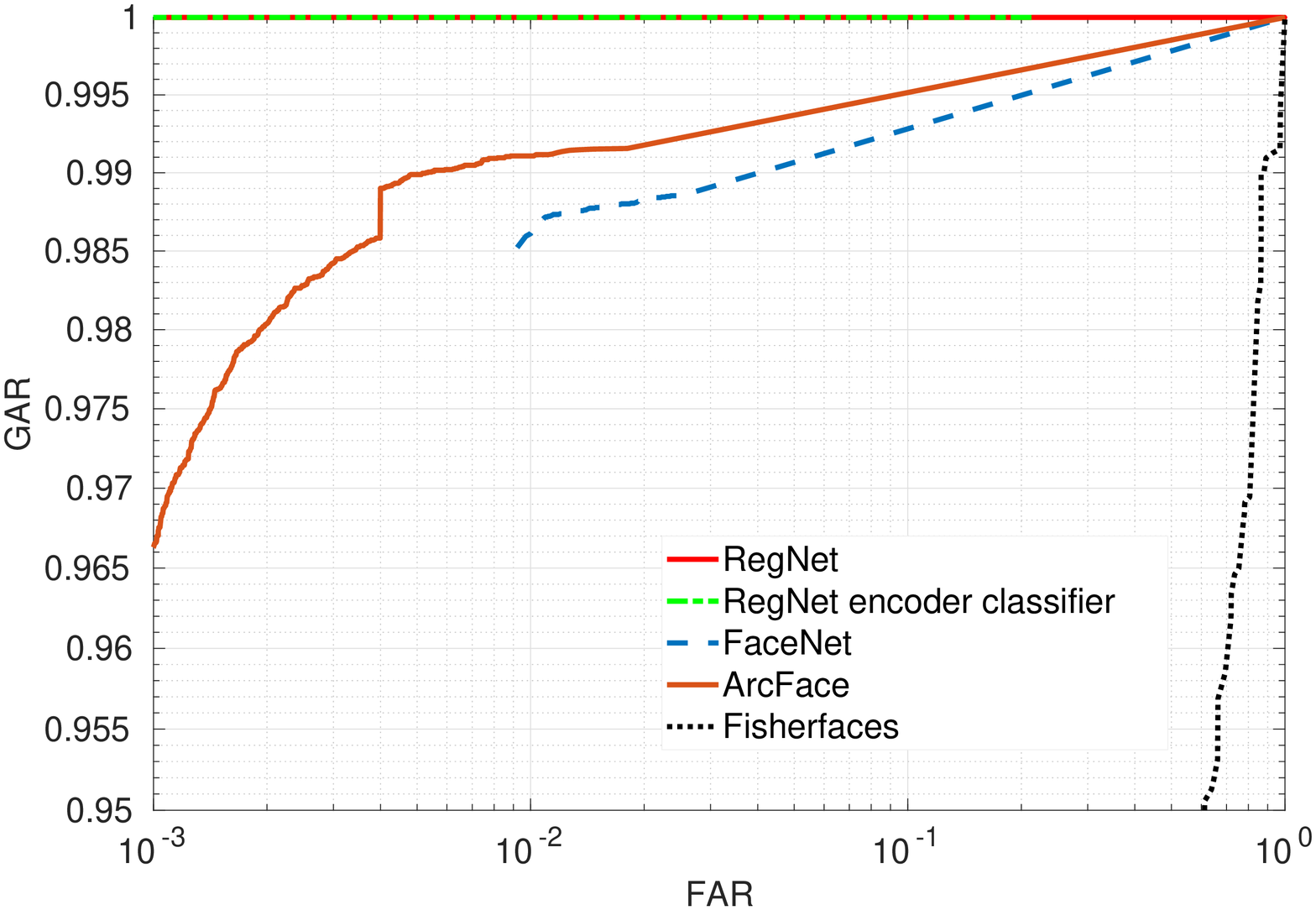}
  \caption[]{Face dataset - Yale B}
  \label{fig:RocComp_faces_yale}
\end{subfigure}
\begin{subfigure}{.32\textwidth}
  \centering
  \includegraphics[width=\linewidth]{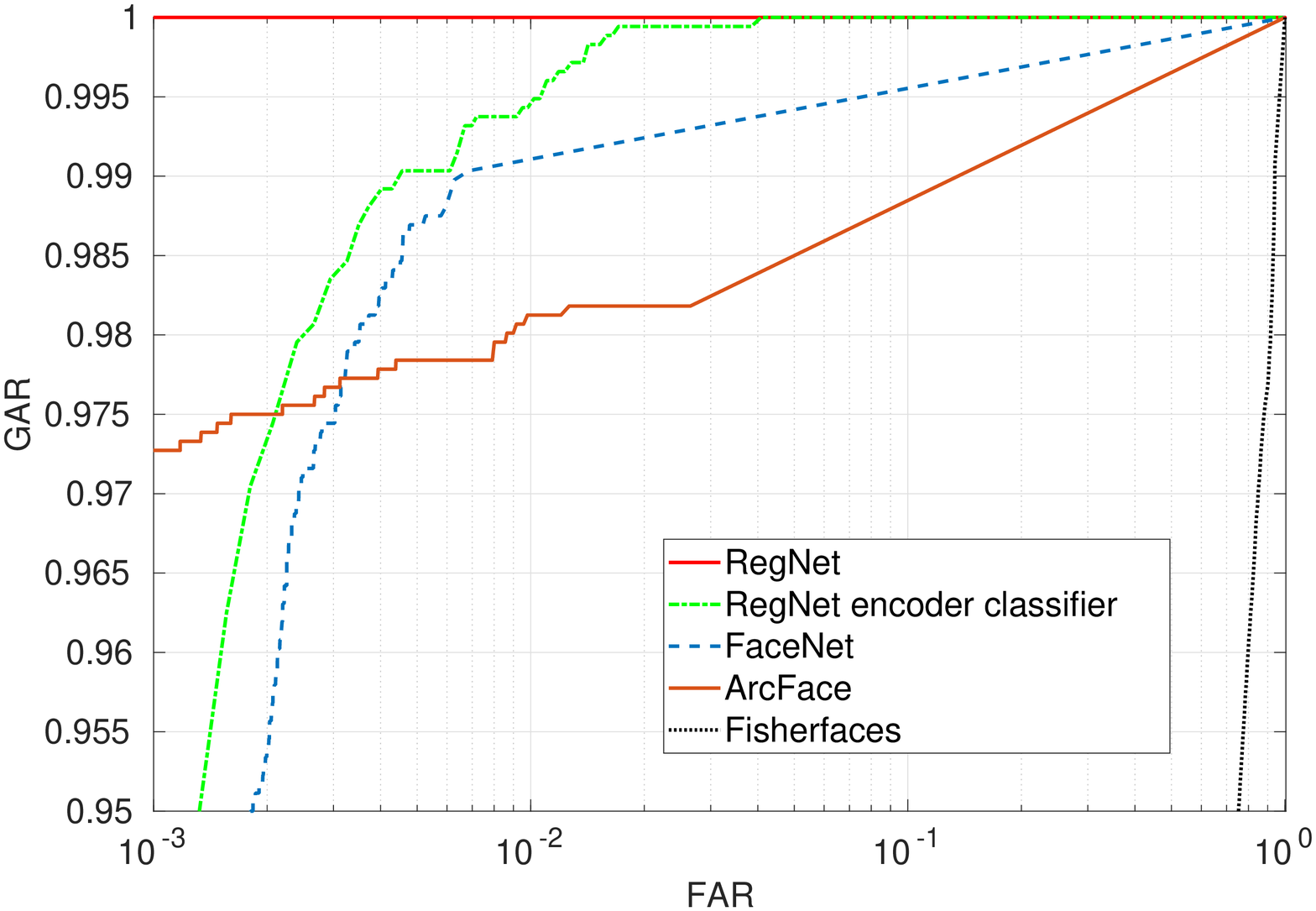}
  \caption[]{Face dataset - MultiPIE }
  \label{fig:RocComp_faces_multipie}
\end{subfigure}
\begin{subfigure}{0.32\textwidth}
  \centering
  \includegraphics[width=\linewidth]{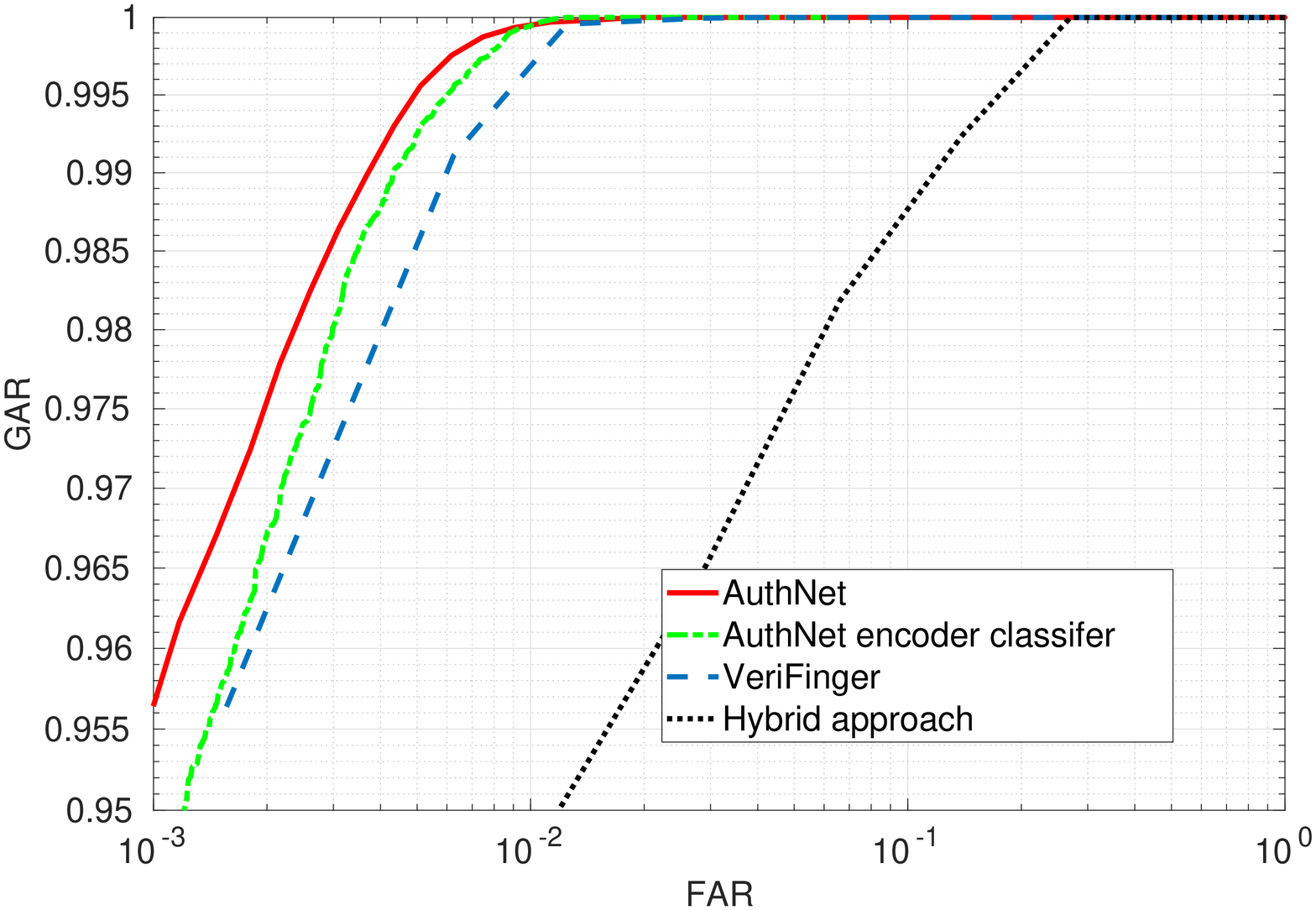}
  \caption[]{{Fingerprint dataset}}
  \label{fig:RocComp_fingerprints}
\end{subfigure}
\vspace*{-0.1cm}
\caption[]{ROC comparison on overall results of $32$ users for faces (a)-(b) and fingerprint (c) datasets. RegNet is compared with the RegNet encoder classifier, FaceNet \cite{schroff2015facenet}, ArcFace \cite{deng2018arcface} and Fisherfaces \cite{swets1996using} in (a)-(b); with RegNet encoder classifier, VeriFinger \cite{verifingerneuro} and the hybrid approach \cite{abraham2011fingerprint} in (c).}
\label{fig:ROC_comparison}
\vspace*{-0.5cm}
\end{figure*}


\subsection{Results}
\textbf{Face authentication.} In this case we compare the results of RegNet with the RegNet encoder classifier, the Fisherfaces approach \cite{swets1996using}, FaceNet \cite{schroff2015facenet} and ArcFace \cite{deng2018arcface}.
The RegNet encoder classifer is a network with the same structure as the RegNet encoder, but trained in a more classical way through sigmoid cross-entropy loss. This network does not employ a variational loss function, therefore it allows us to assess the improvement obtained via the learned mapping with respect to a conventional neural network.
Regarding FaceNet and ArcFace, since it is not possible to train it from scratch because of the extreme data scarcity, we compute the $512$-dimensional embeddings of the input images given a pre-trained network on the CASIA WebFace dataset \cite{yi2014learning}. Then, a classifier is independently trained on the embeddings of each user.

As can be seen from Table \ref{table:EER_faces}, RegNet achieves the highest performance on all the considered metrics. It is important to notice that the RegNet encoder classifier, while sharing the RegNet architecture, achieves lower performance especially at low FAR, see Fig. \ref{fig:ROC_comparison}(a)-(b). This suggests that the well-defined regions in the latent space given by the target distributions yield a more robust classification scheme. 
Indeed, as can be seen in Fig. \ref{fig:hist_faces_multipie}(a) and \ref{fig:hist_faces_yale}(a), RegNet effectively separates authorized and unauthorized users. A good separation is also achieved by other methods, see Fig. \ref{fig:hist_faces_multipie}(b)-(c) and \ref{fig:hist_faces_yale}(b)-(c); however they fail to assign to all the unauthorized users a correct score, yielding some ``leakage'' into the wrong distribution. This behavior can indeed be more clearly noticed in ROC comparison in Fig. \ref{fig:ROC_comparison}(a)-(b). Further, it can also be noticed that the proposed approach performs better at low FAR values when compared to RegNet encoder classifier.

At this point it is also interesting to notice that deep learning based methods are able to achieve higher performance when tested on the Multi-PIE dataset. Even though this dataset is more complex with respect to the Yale B because of the unconstrained acquisitions, it has more samples. For this reason, methods which can learn complex  features from the data will benefit. 
Conversely, traditional approaches such as Fisherfaces by relying on properly aligned and constrained images show a performance drop when tested on the more complex Multi-PIE dataset.

\textbf{Fingerprint authentication.} For fingerprint authentication we compare RegNet with RegNet encoder classifier, Verifinger \cite{verifingerneuro}, and the hybrid approach described in \cite{abraham2011fingerprint}.
As for the equal error rate (EER), the proposed method achieves an EER of $0.476$\% outperforming all the other approaches. In terms of genuine acceptance rate (GAR) at small FAR values, the proposed method outperforms the hybrid approach, and improves over RegNet encoder classifier and Verifinger. As previously observed for the case of face authentication, in Fig. \ref{fig:hist_fingerprints}(a) it can be seen how RegNet effectively separates authorized and unauthorized users.
However, it can also be noticed that the distribution of authorized users spreads more with respect to the case of the face dataset. This might be due to the extremely small number of training samples for the authorized user which is only $10$ prior to the augmentation.
In the case of non deep learning approaches Fig. \ref{fig:hist_fingerprints}(c)-(d), the authorized and unauthorized users do not have a clear scores separation. Additionally, it can be seen that the RegNet encoder classifier is still introducing some ``leakage''.
This aspect can be further noticed in Fig. \ref{fig:ROC_comparison}(c): RegNet outperforms all other methods by achieving highest values of GAR even for low values of FAR.

The above results strongly motivate the intuition behind RegNet: learning the mapping instead of the classification boundaries leads to improved performance and robustness of the classifier.
\section{Conclusions}
We presented a novel strategy to address the biometric authentication problem with deep neural networks. Instead of learning complex boundaries, the proposed approach aims at learning a mapping onto target distributions allowing for simple threshold-based classification.
We demonstrated that RegNet is an effective general-purpose biometric authentication framework which can achieve low EER and good latent space separation as demonstrated through extensive experiments on two different biometric traits. Furthemore, the comparison with a network sharing the same architecture as RegNet but trained in a more standard way, allowed us to show the superiority of the proposed architecture.

\section*{Acknowledgment}
 This work results from the research cooperation with the Sony Technology Center Stuttgart (Sony EuTEC). We would like to thank Sony EuTEC for their feedback and the fruitful discussions.   

\bibliographystyle{IEEEtran}
\bibliography{ms}

\begin{thebibliography}{10}
\providecommand{\url}[1]{#1}
\csname url@samestyle\endcsname
\providecommand{\newblock}{\relax}
\providecommand{\bibinfo}[2]{#2}
\providecommand{\BIBentrySTDinterwordspacing}{\spaceskip=0pt\relax}
\providecommand{\BIBentryALTinterwordstretchfactor}{4}
\providecommand{\BIBentryALTinterwordspacing}{\spaceskip=\fontdimen2\font plus
\BIBentryALTinterwordstretchfactor\fontdimen3\font minus
  \fontdimen4\font\relax}
\providecommand{\BIBforeignlanguage}[2]{{%
\expandafter\ifx\csname l@#1\endcsname\relax
\typeout{** WARNING: IEEEtran.bst: No hyphenation pattern has been}%
\typeout{** loaded for the language `#1'. Using the pattern for}%
\typeout{** the default language instead.}%
\else
\language=\csname l@#1\endcsname
\fi
#2}}
\providecommand{\BIBdecl}{\relax}
\BIBdecl

\bibitem{goodfellow2014generative}
I.~Goodfellow, J.~Pouget-Abadie, M.~Mirza, B.~Xu, D.~Warde-Farley, S.~Ozair,
  A.~Courville, and Y.~Bengio, ``Generative adversarial nets,'' in
  \emph{Advances in neural information processing systems}, 2014, pp.
  2672--2680.

\bibitem{almahairi2018augmented}
A.~Almahairi, S.~Rajeswar, A.~Sordoni, P.~Bachman, and A.~Courville,
  ``Augmented cyclegan: Learning many-to-many mappings from unpaired data,''
  \emph{arXiv preprint arXiv:1802.10151}, 2018.

\bibitem{krizhevsky2012imagenet}
A.~Krizhevsky, I.~Sutskever, and G.~E. Hinton, ``Imagenet classification with
  deep convolutional neural networks,'' in \emph{Advances in neural information
  processing systems}, 2012, pp. 1097--1105.

\bibitem{fawzi2017classification}
A.~Fawzi, S.-M. Moosavi-Dezfooli, P.~Frossard, and S.~Soatto, ``Classification
  regions of deep neural networks,'' \emph{arXiv preprint arXiv:1705.09552},
  2017.

\bibitem{robustnessmagazine2017fawzi}
A.~Fawzi, S.~Moosavi-Dezfooli, and P.~Frossard, ``The robustness of deep
  networks: A geometrical perspective,'' \emph{IEEE Signal Processing
  Magazine}, vol.~34, no.~6, pp. 50--62, Nov 2017.

\bibitem{chaoqiang2004hierarchical}
L.~Chaoqiang, X.~Tao, and L.~Hui, ``A hierarchical hough transform for
  fingerprint matching.''

\bibitem{ratha1996real}
N.~K. Ratha, K.~Karu, S.~Chen, and A.~K. Jain, ``A real-time matching system
  for large fingerprint databases,'' \emph{IEEE Transactions on Pattern
  Analysis \& Machine Intelligence}, no.~8, pp. 799--813, 1996.

\bibitem{sha2004orientation}
L.~Sha and X.~Tang, ``Orientation-improved minutiae for fingerprint matching,''
  in \emph{Pattern Recognition, 2004. ICPR 2004. Proceedings of the 17th
  International Conference on}, vol.~4.\hskip 1em plus 0.5em minus 0.4em\relax
  IEEE, 2004, pp. 432--435.

\bibitem{abraham2011fingerprint}
J.~Abraham, P.~Kwan, and J.~Gao, ``Fingerprint matching using a hybrid shape
  and orientation descriptor,'' in \emph{State of the art in Biometrics}.\hskip
  1em plus 0.5em minus 0.4em\relax InTech, 2011.

\bibitem{jiang2016direct}
L.~Jiang, T.~Zhao, C.~Bai, A.~Yong, and M.~Wu, ``A direct fingerprint minutiae
  extraction approach based on convolutional neural networks,'' in \emph{Neural
  Networks (IJCNN), 2016 International Joint Conference on}.\hskip 1em plus
  0.5em minus 0.4em\relax IEEE, 2016, pp. 571--578.

\bibitem{swets1996using}
D.~L. Swets and J.~J. Weng, ``Using discriminant eigenfeatures for image
  retrieval,'' \emph{IEEE Transactions on Pattern Analysis \& Machine
  Intelligence}, no.~8, pp. 831--836, 1996.

\bibitem{deng2012extended}
W.~Deng, J.~Hu, and J.~Guo, ``Extended src: Undersampled face recognition via
  intraclass variant dictionary,'' \emph{IEEE Trans. Pattern Anal. Mach.
  Intell.}, vol.~34, no.~9, pp. 1864--1870, 2012.

\bibitem{he2005face}
X.~He, S.~Yan, Y.~Hu, P.~Niyogi, and H.-J. Zhang, ``Face recognition using
  laplacianfaces,'' \emph{IEEE transactions on pattern analysis and machine
  intelligence}, vol.~27, no.~3, pp. 328--340, 2005.

\bibitem{schroff2015facenet}
F.~Schroff, D.~Kalenichenko, and J.~Philbin, ``Facenet: A unified embedding for
  face recognition and clustering,'' in \emph{Proceedings of the IEEE
  conference on computer vision and pattern recognition}, 2015, pp. 815--823.

\bibitem{taigman2014deepface}
Y.~Taigman, M.~Yang, M.~Ranzato, and L.~Wolf, ``Deepface: Closing the gap to
  human-level performance in face verification,'' in \emph{Proceedings of the
  IEEE conference on computer vision and pattern recognition}, 2014, pp.
  1701--1708.

\bibitem{deng2018arcface}
J.~Deng, J.~Guo, N.~Xue, and S.~Zafeiriou, ``Arcface: Additive angular margin
  loss for deep face recognition,'' \emph{arXiv preprint arXiv:1801.07698},
  2018.

\bibitem{makhzani2015adversarial}
A.~Makhzani, J.~Shlens, N.~Jaitly, I.~Goodfellow, and B.~Frey, ``Adversarial
  autoencoders,'' \emph{arXiv preprint arXiv:1511.05644}, 2015.

\bibitem{kingma2013auto}
D.~P. Kingma and M.~Welling, ``Auto-encoding variational bayes,'' \emph{arXiv
  preprint arXiv:1312.6114}, 2013.

\bibitem{resnet}
K.~He, X.~Zhang, S.~Ren, and J.~Sun, ``Deep residual learning for image
  recognition,'' in \emph{Proceedings of the IEEE conference on computer vision
  and pattern recognition}, 2016, pp. 770--778.

\bibitem{gross2010multi}
R.~Gross, I.~Matthews, J.~Cohn, T.~Kanade, and S.~Baker, ``Multi-pie,''
  \emph{Image and Vision Computing}, vol.~28, no.~5, pp. 807--813, 2010.

\bibitem{zhang2017mixup}
H.~Zhang, M.~Cisse, Y.~N. Dauphin, and D.~Lopez-Paz, ``mixup: Beyond empirical
  risk minimization,'' \emph{arXiv preprint arXiv:1710.09412}, 2017.

\bibitem{georghiades2001few}
A.~S. Georghiades, P.~N. Belhumeur, and D.~J. Kriegman, ``From few to many:
  Illumination cone models for face recognition under variable lighting and
  pose,'' \emph{IEEE transactions on pattern analysis and machine
  intelligence}, vol.~23, no.~6, pp. 643--660, 2001.

\bibitem{cappelli2007fingerprint}
R.~Cappelli, M.~Ferrara, A.~Franco, and D.~Maltoni, ``Fingerprint verification
  competition 2006,'' \emph{Biometric Technology Today}, vol.~15, no. 7-8, pp.
  7--9, 2007.

\bibitem{verifingerneuro}
S.~VeriFinger, ``Neuro technology (2010),'' \emph{VeriFinger, SDK Neuro
  Technology}.

\bibitem{yi2014learning}
D.~Yi, Z.~Lei, S.~Liao, and S.~Z. Li, ``Learning face representation from
  scratch,'' \emph{arXiv:1411.7923}, 2014.

\end{thebibliography}
\end{document}